\documentclass[10pt,twocolumn,letterpaper]{article}

\usepackage[pagenumbers]{cvpr} 

\newcommand{\parag}[1]{\vspace{0.5em}\noindent\textbf{#1.}}

\newcommand{\archname}{{\textit{InstAlign}}\xspace}

\definecolor{cvprblue}{rgb}{0.21,0.49,0.74}
\usepackage[pagebackref,breaklinks,colorlinks,allcolors=cvprblue]{hyperref}
\usepackage{graphicx}
\usepackage{caption}
\usepackage{subcaption}
\usepackage{booktabs}
\usepackage{colortbl}
\usepackage{multirow}
\usepackage{array}
\usepackage{xcolor}
\usepackage{xspace}
\usepackage{placeins}
\usepackage{amsmath,amssymb}
\usepackage{algorithm}
\usepackage{algpseudocode}
\usepackage{wrapfig}
\usepackage[capitalize,noabbrev]{cleveref}

\def\paperID{1} 
\def\confName{CVPR}
\def\confYear{2026}

\title{Phrase-Instance Alignment for Generalized Referring Segmentation}

\author{E-Ro Nguyen$^1$ \quad Hieu Le$^2$ \quad Dimitris Samaras$^1$ \quad Michael S. Ryoo$^1$ \\ $^1$Stony Brook University \quad $^2$UNC Charlotte \\
\href{https://eronguyen.github.io/InstAlign}{\texttt{eronguyen.github.io/InstAlign}
}}

\begin{document}

\maketitle
\begin{abstract}
Generalized Referring expressions can describe one object, several related objects, or none at all. Existing generalized referring segmentation (GRES) models treat all cases alike—predicting a single binary mask and ignoring how linguistic phrases correspond to distinct visual instances. To this end, we reformulate GRES as an instance-level reasoning problem, where the model first predicts multiple instance-aware object queries conditioned on the referring expression, then aligns each with its most relevant phrase. This alignment is enforced by a Phrase–Object Alignment (POA) loss that builds fine-grained correspondence between linguistic phrases and visual instances. Given these aligned object instance queries and their learned relevance scores, the final segmentation and the no-target case are both inferred through a unified relevance-weighted aggregation mechanism.
This instance-aware formulation enables explicit phrase–instance grounding, interpretable reasoning, and robust handling of complex or null expressions. Extensive experiments on the gRefCOCO and Ref-ZOM benchmarks demonstrate that our method significantly advances state-of-the-art performance by $3.22\%$ cIoU and $12.25\%$ N-acc.
\end{abstract}

\begin{figure}
    \centering
    \includegraphics[width=\linewidth]{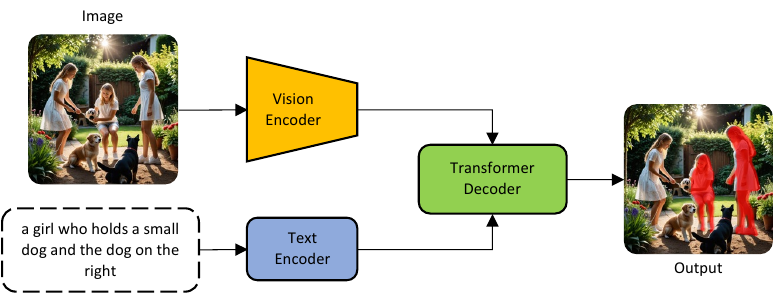}
    \makebox[\linewidth]{(a) Region-based approach.}
    \includegraphics[width=\linewidth]{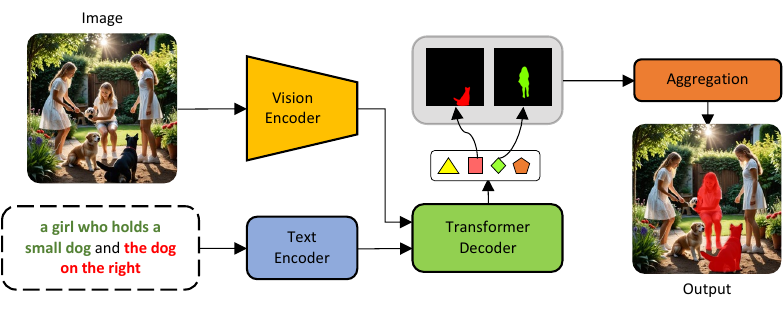}
     \makebox[\linewidth]{(b) Instance-aware approach (Ours).}
    \caption{
    Comparison between previous region-based GRES methods and our instance-aware approach. 
    (a) Prior works predict a single foreground mask directly from fused image-text features, often failing to distinguish between multiple referred instances. 
    (b) Our method explicitly segments individual object instances, aligns them with corresponding phrases in the text, and aggregates those instances to form the final output mask.
    }
    
    \label{fig:teaser}
    \vspace{-5mm}
\end{figure}

\section{Introduction}
\label{sec:intro}

While classic Referring Expression Segmentation (RES)~\cite{mao_generation_2016, ding_vision-language_2021, wang_cris_2022, huang_referring_2020} focuses on single-object descriptions, Generalized Referring Segmentation (GRES) extends to more realistic scenarios where an expression may describe multiple related objects or none at all—“two men on the left,” “all the cars in the lot,” “the elephant on the couch.” This broader setting demands models that can reason about how different parts of a sentence correspond to different visual entities. Yet, most current GRES approaches~\cite{liu_gres_2023,shah_lqmformer_2024, hu_beyond_2023, luo_hdc_2024} still predict a single foreground mask for the entire expression, effectively flattening this linguistic structure into one undifferentiated region. Without explicitly reasoning over phrase–object relationships, they struggle to separate related instances, capture attributes, or generalize to compositional and relational language. As shown in Fig.~\ref{fig:teaser}, existing models often over-segment or confuse closely related instances. As the linguistic complexity of prompts increases, so does the need for more structured reasoning. 

To address this, we propose \archname, a novel framework for phrase–instance reasoning in GRES.
Instead of predicting a single merged region, \archname first performs phrase-conditioned instance segmentation: the model predicts multiple instance masks and learns to align each with its most relevant phrase in the input expression with a relevance score. The final segmentation is then obtained by aggregating those instances. This formulation allows structured reasoning over the correspondence between language phrases and visual objects, leading to more precise localization, greater compositional generalization, and better interpretability.

To do so, we introduces two key ideas. (1) We extend a query-based instance segmentation backbone to operate under referring expressions, enabling the model to produce language-conditioned object queries that each correspond to a distinct, interpretable instance. This instance-aware formulation introduces explicit object-level supervision, which has not been done in GRES. (2) We propose a Phrase–Object Alignment (POA) loss that explicitly links each object query to its most relevant linguistic phrase. Unlike the implicit cross-modal attention used in previous works, POA provides direct alignment supervision, ensuring fine-grained semantic consistency between visual instances and textual descriptions.
Given these object queries and their relevance scores, we formulate GRES as a problem of relevance-weighted instance aggregation. We design an Instance Aggregation (IA) module that fuses the predictions of all object queries according to their learned relevance weights, allowing the model to selectively merge relevant instances while suppressing distractors. Similarly, a no-target predictor uses the same relevance representation to detect cases where no valid object is referred via a simple classifier head.

We evaluate \archname~on two widely used benchmarks for GRES: gRefCOCO~\cite{liu_gres_2023} and Ref-ZOM~\cite{hu_beyond_2023}. Our experiments show that \archname~significantly outperforms state-of-the-art GRES methods(e.g., $+3.22\%$ cIoU and $+12.25\%$ N-acc in gRefCOCO). The results highlight \archname's ability to handle complex referring expressions, setting a new standard in the field. Our main contributions are:

\begin{itemize} 
    \item We introduces explicit instance-level supervision into GRES for the first time, allowing each object query to specialize into a distinct, interpretable instance tied to the referring expression.
    \item We propose POA, a differentiable alignment mechanism that links each instance to its most relevant phrase within the expression. Unlike prior GRES models that rely on implicit cross-modal attention, POA provides direct supervision of phrase–instance correspondence, yielding semantically grounded and human-interpretable object queries.
    \item Leveraging instance relevance scores, we propose Relevance-Weighted Aggregation mechanism to dynamically merge relevant instances and improve robustness in distinguishing no-target scenarios.
    
\end{itemize}

\section{Related Work}

\paragraph{Referring Expression Segmentation (RES).}
RES focuses on segmenting regions in an image described by a referring expression. Early works~\cite{mao_generation_2016, huang_referring_2020, wang_cris_2022} rely on convolutional architectures, but recent models increasingly adopt Transformer-based backbones for better vision-language alignment~\cite{ding_vision-language_2021, xu_bridging_2023, su_language_2023, yang_lavt_2022, chng_mask_2024, nguyen-truong_vision-aware_2024, MeViS, MeViSv2, ding2025multimodalreferringsegmentationsurvey, ding2026grexgeneralizedreferringexpression}. For instance, LAVT~\cite{yang_lavt_2022} shows the benefits of early cross-modal fusion, and CRIS~\cite{wang_cris_2022} leverages CLIP for contrastive pixel-text learning. Recent approaches like ReMamber~\cite{yang_remamber_2024} and Mask Grounding~\cite{chng_mask_2024} explore more explicit image-text token correspondence and memory-based interaction.

\paragraph{Generalized RES (GRES).}
GRES extends RES to handle expressions referring to multiple or zero objects. ReLA~\cite{liu_gres_2023} introduces a region-based GRES baseline using relational attention between image regions and text tokens. DMMI~\cite{hu_beyond_2023} uses dual decoders to reconstruct entity phrases and enhance alignment. LQMFormer~\cite{shah_lqmformer_2024} introduces dynamic query fusion and auxiliary objectives to mitigate query collapse. Although these models adopt query-based architectures, they supervise only the aggregated segmentation output, not the individual queries themselves. As a result, the query slots often remain entangled and lack object-level specialization, which limits interpretability and performance on compositional queries. MABP~\cite{li_bring_2024} integrates linguistic features into initialized queries but only supervises each query with a fixed patch of the image. In contrast, our method directly supervises each object query to represent a referred instance and introduces a novel Phrase-Object Alignment (POA) loss to ground each query in specific linguistic phrases. This enables interpretable and structured reasoning that previous GRES methods cannot support.

\paragraph{LLM-based Vision-Language Models.}
Large Vision-Language Models (VLMs) such as LISA~\cite{lai_lisa_2024}, GSVA~\cite{xia_gsva_2024}, and VistaLLM~\cite{vistallm} have shown strong performance on vision-language tasks, including segmentation. These models leverage large-scale pretraining and introduce modules such as learned SEG tokens or contour-aware grounding. Others, such as SAM4MLLM~\cite{chen_sam4mllm_2024} and PSALM~\cite{zhang2024psalmp}, integrate promptable segmentation with LLM-derived semantics.

While promising, current LLM-based approaches are computationally intensive and lack the task-specific supervision needed for fine-grained, phrase-aware segmentation in GRES. Our approach is complementary, offering a lightweight, trainable framework that makes full use of available supervision for structured instance-level grounding.

\begin{figure*}[ht!]
    \centering
    \includegraphics[width=\textwidth]{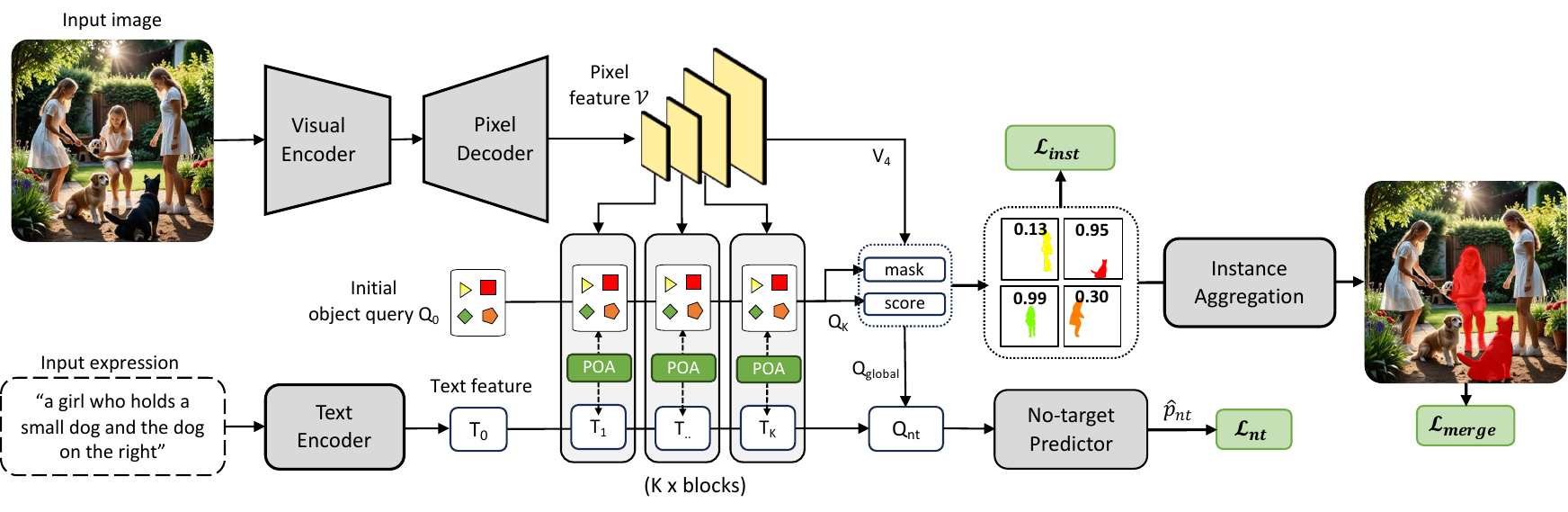}


    \caption{\textbf{Overview of \archname.} The model takes an image and the corresponding referring expression as input. After the encoding process, a set of learnable object queries is refined through transformer decoder blocks to enrich the visual and text cues. During training, object queries are supervised to match referred instances and are semantically aligned with referring phrases via the Instance Supervision $\mathcal{L}_{inst}$  and Phrase-Object Alignment (POA) loss. Meanwhile, instance masks are weighted and aggregated using Instance Aggregation to produce the final segmentation and a no-target predictor takes both queries and text features to handle the no-target scenario.}
    \vspace{-5mm}
    \label{fig:overview}
\end{figure*}

\section{Method}

\archname operates with two main functionalities: (1) phrase-conditioned instance segmentation, which predicts multiple object queries with associated relevance scores, and (2) relevance-weighted aggregation, which fuses all instance predictions into a final output by weighting each according to its learned relevance to the expression.

More specifically, we first extend an instance-aware segmentation framework to the GRES setting (\cref{sec:supervision}), enabling explicit object-level supervision. Next, we introduce the Phrase–Object Alignment (POA) loss (\cref{sec:alignment}) to align each object query with its most relevant linguistic phrase, reinforcing semantic consistency between visual instances and text. Finally, we formulate GRES as a relevance-weighted aggregation problem (\cref{sec:merging}), where all instance predictions are softly merged according to their learned relevance scores to produce the final segmentation.
\Cref{fig:overview} illustrates the overall architecture.
The model starts from a set of randomly initialized object queries that iteratively attend to multi-scale visual features and the encoded referring expression, evolving into distinct, text-aware instance representations.


\subsection{Instance-aware Segmentation Framework}
\label{sec:supervision}
Most prior GRES methods collapse all referred regions into a single mask, leaving object queries entangled and semantically vague. We instead start from an instance segmentation backbone Mask2Former ~\cite{cheng_masked-attention_2022} and condition it on the input text. By adopting this instance-aware segmentation framework, we reformulate the GRES problem from an end-to-end single-mask prediction into a structured instance-level reasoning process, where the model first discovers candidate object instances conditioned on language and before merging them to the final prediction.

First, each image is encoded into multi-scale pixel features $\mathcal{V} = \{V_i \in \mathbb{R}^{C \times H_i \times W_i}\}_{i = 1}^{4}$ using a visual encoder and pixel decoder, while the referring expression is encoded via a BERT-based encoder~\cite{liu_roberta_2019} into token features $T_0 \in \mathbb{R}^{L \times C}$.  Here, $H_i$ and $W_i$, $C$, $L$ denote the height and width of the feature maps in the $i$-th stage, the channel dimension, and the length of the expression, respectively. These features interact with a set of $N$ learnable object queries $Q_0 \in \mathbb{R}^{N \times C}$ through $K$ transformer decoder layers.

Each decoder block performs bi-directional cross-attention~\cite{liu_gres_2023, nguyen-truong_vision-aware_2024} between queries, visual features $V_k$, and text features $T_{k-1}$, producing refined queries $Q_k$ and updated text $T_k$. The final queries $Q_K$ are used to predict $N$ instance masks $\hat{s} \in \mathbb{R}^{\frac{H}{4} \times \frac{W}{4} \times N}$ and relevance scores $\hat{p} \in [0, 1]^N$.

Following ~\cite{cheng_masked-attention_2022}, we supervise object queries using instance-level annotations. Given $M$ ground-truth masks $\{s_i\}_{i=1}^M$, we compute optimal assignments via Hungarian matching~\cite{kuhn_hungarian_1955}. The per-instance matching cost for  $i$-th predicted instance and $j$-th ground-truth instance is defined as:
\begin{equation}
\label{eq:matching}
\mathcal{L}_{\text{match}}(i, j) = \lambda_{\text{score}} \mathcal{L}_{\text{score}}(\hat{p}_i, 1) + \lambda_{\text{mask}} \mathcal{L}_{\text{mask}}(\hat{s}_i, s_j) 
\end{equation}
Matched queries are trained with $\mathcal{L}_{\text{match}}$, and unmatched ones are trained with $\mathcal{L}_{\text{score}}(\hat{p}_i, 0)$. This forms the instance loss $\mathcal{L}_{\text{inst}}$. In essence, $\mathcal{L}_{\text{inst}}$ encourages each object query to specialize into a distinct, spatially grounded instance while remaining consistent with the linguistic structure of the expression, forming the foundation for later phrase–object alignment and aggregation.
Details of the architecture are provided in the Supplementary Materials.


\subsection{Phrase-Object Alignment}
\label{sec:alignment}

\begin{figure}[t!]
    \centering
    \includegraphics[width=\linewidth]{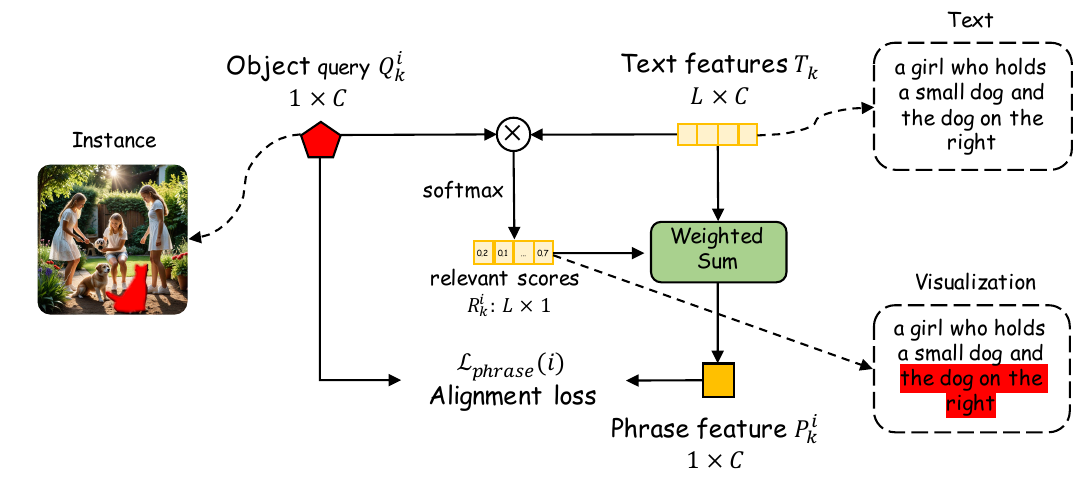}
    \caption{\textbf{POA loss.} The POA objective aligns each object query with its most relevant phrase in the referring expression, establishing fine-grained correspondence between linguistic components and visual instances. Dashed arrows are shown for illustration only.}
    \label{fig:alignment}
\end{figure}

Instance-aware segmentation provides the structural decomposition of the input image into object segments.
Next, we complement it with a soft association between each object segment and the phrases in the referring expression, enabling the model to reason about how linguistic components map to individual visual instances.
This association forms the basis of our Phrase–Object Alignment (POA) loss (see \cref{fig:alignment}), which strengthens the semantic correspondence between visual queries and textual phrases, ensuring that each predicted instance is linguistically grounded in the expression according to its learned relevance probability.

Specifically, POA proceeds in three steps:
\begin{enumerate}
    \item \textbf{Token-to-query relevance.} For each word token, we measure how strongly every object query attends to it, forming a phrase–object attention map.
    \item \textbf{Phrase embedding construction.} Using this attention map, we compute a soft linguistic embedding for each object query that summarizes the textual context most related to the corresponding object segment.
    \item \textbf{Cross-modal alignment.} We then align each object query with its phrase embedding through a similarity loss, enforcing consistent visual–linguistic grounding.
\end{enumerate}


\vspace{2mm}
More specifically, after each decoder block, we obtain refined object queries $Q_k \in \mathbb{R}^{N \times C}$ and text features $T_k \in \mathbb{R}^{L \times C}$. We compute a phrase-object relevance matrix $R_k \in [0, 1]^{N \times L}$ using scaled dot-product attention:
\begin{equation}
    R_k = \text{softmax}\left( \frac{Q_k T_k^\top}{\sqrt{C}} \right),
\end{equation}
where each row of $R_k$ captures how strongly a given object query attends to each word in the referring expression. Using this relevance matrix $R_k$, we derive a soft phrase embedding $P_k$ for each object query by weighted summation:
\begin{equation}
    P_k = R_k T_k
\end{equation}
We then compute a cosine similarity between each query and its corresponding phrase embedding and define the POA loss as:
\begin{equation}
    \mathcal{L}_{\text{phrase}}(i) = 1 - \text{sim}(Q_k^i, P_k^i),
\end{equation}

which encourages each object query embedding to carry semantic information that aligns with the text phrases it corresponds to.  It is particularly effective for disambiguating similar objects (e.g., two dogs) or resolving expressions with compositional structure (e.g., attributes and relations). In essence, each object query serves as a unified representation that bridges the visual and linguistic modalities: it can be used both to anchor the query to its corresponding phrase in the expression and to predict the instance mask of the referred object.

We then extend the instance matching cost $\mathcal{L}_{match}$ (Eq.\ref{eq:matching}) by adding \( \mathcal{L}_{{phrase}}\), weighted by $\lambda_{{phrase}}$ to guide predictions toward both spatial and linguistic relevance. This additional term encourages stronger semantic grounding by ensuring that each predicted instance is not only spatially accurate but also linguistically relevant. 

\begin{figure}
    \centering
    \includegraphics[width=0.48\textwidth,trim=0 0 0 0,clip]{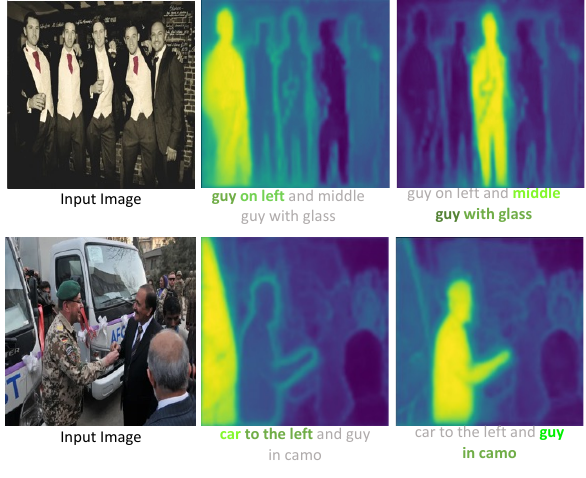}
    \vspace{-8mm}
    \caption{\textbf{Phrase-Query Alignment.} Segmentation heatmaps of two queries and their corresponding soft-phrases. We highlight words with highest weights (the exact weights are in the supplementary material). In each case, the highlighted words aligns with the associated instance.}
    \label{fig:alignment_visualization}
\end{figure}

Fig.~\ref{fig:alignment_visualization} visualizes instance heatmaps from high-relevant queries on two image examples and their corresponding phrases in the input text. The relevant parts of the referring expression, as indicated by high phrase-object relevance scores, are highlighted in yellow. These phrase-object associations are both meaningful and consistent with human intuition, especially in multi-object scenarios. This demonstrates that POA effectively encourages each object query to focus on and claim the linguistic phrase that best matches the visual instance it segments. 

\subsection{Relevance-Weighted Aggregation}
\label{sec:merging}
Having linked each object query to its most relevant phrase through POA, we now leverage these instance-aware representations to infer the final output in a relevance-weighted manner.
We design two lightweight modules built on this shared formulation: one that aggregates instance masks when the referred objects are present, and another that predicts a no-target output when none exist.
Both modules operate directly on the instance-aware queries and use their learned relevance probabilities as weighting factors, providing a unified and efficient inference process for all expression types.

\begin{figure}[!t]
    \centering
    \includegraphics[width=\linewidth,trim=0 20 0 0,clip]{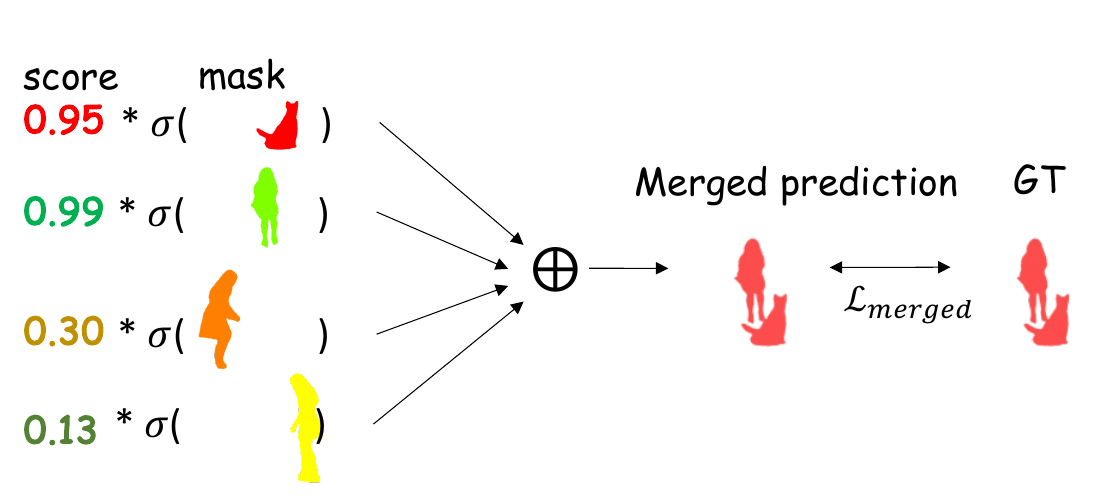}
    \caption{\textbf{Instance Aggregation.}  All instance masks are softly merged according to their learned relevance scores. $\sigma$ denotes the PReLU activation that acts as a learnable dynamic threshold to suppress background noise.}
    \label{fig:adaptive_merging}
    \vspace{-5mm}
\end{figure}


\parag{Instance Aggregation (IA)} Given predicted instance masks 
$\{\hat{s}_i\}$ and their phrase–object relevance probabilities 
$\{\hat{p}_i\}$, we compute the final segmentation mask as a relevance-weighted mixture of all instances:


\begin{equation}
    \mathcal{M}_{merged} = \text{Sigmoid}\left(\sum_{i = 1}^{N}{(\hat{p}_i \cdot \sigma(\hat{s}_i))}\right),
\end{equation}
where $\sigma(\cdot)$ denotes a PReLU activation that suppresses background noise and stabilizes training. 

Since the predicted mask logits range from $[-\infty,\infty]$, PReLU acts as a learnable dynamic threshold, adaptively suppressing background noise while preserving confident foreground regions. This formulation softly merges all candidate instances, emphasizing those that are semantically aligned with the expression while attenuating distractors.
Unlike hard selection strategies, this continuous weighting is fully differentiable and allows the model to handle multi-object and compositional expressions gracefully. We provide ablation study for this design in \cref{tab:ablation_full}.

We supervise the merging process via a mask loss between the final merged mask $\mathcal{M}_{merged}$ and ground-truth merged mask $\mathcal{M}_{\text{gt}}$ to encourage the model to learn both accurate instance segmentation and effective weighting for aggregation.
\begin{equation}
    \mathcal{L}_{merged} = \mathcal{L}_{mask}(\mathcal{M}_{merged}, \mathcal{M}_{gt})
\end{equation}

\parag{No-target Predictor}
\label{sec:no_target}
The no-target scenario in GRES arises when the input expression does not refer to any object in the image. Starting from the learned relevance probabilities $\{\hat{p}_i\}$ obtained through POA, we introduce a lightweight no-target prediction module to determine whether the referring expression corresponds to any valid object in the image.
When all relevance scores are weak, the model should infer a no-target case and vice versa. To make this decision, we aggregate the instance-aware object queries $\{Q\}$ according to their relevance probabilities $\{\hat{p}\}$:



\vspace{-3mm}
\begin{equation}
Q_{global} = \sum_{i = 1}^{N}{(\hat{p}_i \cdot Q^i)},
\vspace{-2mm}
\end{equation}
yielding a compact global feature $Q_{global}$ that summarizes the overall alignment between language and vision based on the extracted object queries and their relevance scores. We then concatenate this feature with a sentence-level text embedding $T_{sen} = \text{Average}(T_K, \text{dim = 0}) $ and feed the result to a small MLP classifier:

\begin{align*}
        T_{sen} = & \text{Average}(T_K, \text{dim = 0}) \\
        Q_{nt} = &\text{Concat}(Q_{global}, T_{sen}) \\
        \hat{p}_{nt} = &\text{MLP}(Q_{nt}),  \\   
\end{align*}
which outputs a single binary probability indicating the absence of referred objects. Given our relevance-weighted instance-aware representation, this predictor can be viewed as a simple learnable aggregation—conceptually analogous to the final mask inference but operating in the binary domain. As shown in \cref{tab:ablation_full}, this unified design yields accurate and robust no-target detection without introducing additional complexity.

\subsection{Training Objectives}
We supervise our framework with three training objectives: Our proposed instance supervision $\mathcal{L}_{inst}$, which guides the association between object queries, targeted object masks, and relevant input text phrases (section \ref{sec:supervision}); and two losses are adopted from ReLA~\cite{liu_gres_2023}: $\mathcal{L}_{merged}$ and $\mathcal{L}_{nt}$ which are BCE loss to supervise the final mask loss and measure the existence of instances from $\hat{p}_{nt}$, respectively. Therefore, the total loss can be formulated as:
\begin{equation}
    \begin{split}
        \mathcal{L}_{total} = \lambda_{merged}\mathcal{L}_{merged} + \lambda_{inst}\mathcal{L}_{inst}
        + \lambda_{nt}\mathcal{L}_{nt},
    \end{split}
\end{equation}
where $\lambda_{merged}, \lambda_{inst}, \lambda_{nt}$ are the scalar coefficients. 


\begin{table*}[t!]
\centering
\resizebox{0.97\textwidth}{!}{%
\begin{tabular}{@{}lcccccccccc@{}}
\toprule
\multicolumn{1}{l|}{}                                     & \multicolumn{1}{c|}{}                                    & \multicolumn{3}{c|}{\textbf{val}}                                  & \multicolumn{3}{c|}{\textbf{testA}}                                & \multicolumn{3}{c}{\textbf{testB}} \\ \cmidrule(l){3-11} 
\multicolumn{1}{l|}{\multirow{-2}{*}{\textbf{Method}}}    & \multicolumn{1}{c|}{\multirow{-2}{*}{\textbf{Backbone}}} & cIoU  & gIoU  & \multicolumn{1}{c|}{N-acc}                         & cIoU  & gIoU  & \multicolumn{1}{c|}{N-acc}                         & cIoU       & gIoU      & N-acc     \\ \midrule

\multicolumn{11}{c}{\textit{RES   Methods}}                                                                                                                                                                                                                                                         \\ \midrule
\multicolumn{1}{l|}{MattNet~\cite{yu_mattnet_2018}}                              & \multicolumn{1}{c|}{ResNet-101}                          & 47.51 & 48.24 & \multicolumn{1}{c|}{41.15}                         & 58.66 & 59.30 & \multicolumn{1}{c|}{44.04}                         & 45.33      & 46.14     & 41.32     \\
\multicolumn{1}{l|}{VLT~\cite{ding_vision-language_2021}}                                  & \multicolumn{1}{c|}{DarkNet-53}                          & 52.61 & 52.00 & \multicolumn{1}{c|}{47.17}                         & 62.19 & 63.20 & \multicolumn{1}{c|}{48.74}                         & 50.52      & 50.88     & 48.46     \\
\multicolumn{1}{l|}{CRIS~\cite{wang_cris_2022}}                                 & \multicolumn{1}{c|}{ResNet-101}                          & 55.34 & 56.27 & \multicolumn{1}{c|}{-}                             & 63.82 & 63.42 & \multicolumn{1}{c|}{-}                             & 51.04      & 51.79     & -         \\
\multicolumn{1}{l|}{LAVT~\cite{yang_lavt_2022}}                                 & \multicolumn{1}{c|}{Swin-B}                              & 57.64 & 58.40 & \multicolumn{1}{c|}{49.32}                         & 65.32 & 65.90 & \multicolumn{1}{c|}{49.25}                         & 55.04      & 55.83     & 48.46     \\
\midrule

\multicolumn{11}{c}{\textit{LLM-based   Methods}}                                                                                                                                                                                                                                                   \\ \midrule
\multicolumn{1}{l|}{LISA-7B\textsuperscript{\textdagger}~\cite{lai_lisa_2024}}                              & \multicolumn{1}{c|}{SAM-ViT-H}                           
& 61.63 & 61.76 & \multicolumn{1}{c|}{54.67}                         & 68.5  & 66.27 & \multicolumn{1}{c|}{50.01}                         & 60.63      & 58.84     & 51.91     \\
\multicolumn{1}{l|}{GSVA-7B\textsuperscript{\textdagger}~\cite{xia_gsva_2024}}                              & \multicolumn{1}{c|}{SAM-ViT-H}                           
& 63.29 & 66.47 & \multicolumn{1}{c|}{62.43}                         & 69.93 & 71.08 & \multicolumn{1}{c|}{65.31}                         & 60.47      & 62.23     & 60.56     \\ 
\multicolumn{1}{l|}{VistaLLM-7B\textsuperscript{\textdagger}~\cite{vistallm}}                              & \multicolumn{1}{c|}{EVA-CLIP}                           
& - & 64.4 & \multicolumn{1}{c|}{68.8}                         & - & - & \multicolumn{1}{c|}{-}                         & -      & -     & -     \\ 
\multicolumn{1}{l|}{SAM4MLLM-7B\textsuperscript{\textdagger}~\cite{chen_sam4mllm_2024}}                              & \multicolumn{1}{c|}{SAM-EffViT-XL1}                           
& 65.66 & 68.37 & \multicolumn{1}{c|}{63.71}                         & 69.62 & 69.05 & \multicolumn{1}{c|}{65.96}                         & 62.35      & 63.71     & 61.25     \\ 
\multicolumn{1}{l|}{SAM4MLLM-8B\textsuperscript{\textdagger}~\cite{chen_sam4mllm_2024}}                              & \multicolumn{1}{c|}{LLaVA1.6}                           
& 67.83 & 71.86 & \multicolumn{1}{c|}{66.08}                         & 72.22 & 74.15 & \multicolumn{1}{c|}{63.92}                         & 63.42      & 65.29     & 59.99     \\

\midrule

\multicolumn{11}{c}{\textit{GRES   Methods}}                                                                                                                                                                                                                                                        \\ \midrule

\multicolumn{1}{l|}{ReLA~\cite{liu_gres_2023}}                                 & \multicolumn{1}{c|}{Swin-B}                              
& 62.42 & 63.60 & \multicolumn{1}{c|}{56.37}                         & 69.26 & 70.03 & \multicolumn{1}{c|}{59.02}                         & 59.88      & 61.02     & 58.40     \\
\multicolumn{1}{l|}{LQMFormer~\cite{shah_lqmformer_2024}}                                  & \multicolumn{1}{c|}{Swin-B}                              
& 64.98 & 70.94 & \multicolumn{1}{c|}{67.47}                         & - & - & \multicolumn{1}{c|}{-}                         & -      & -     & -     \\
\multicolumn{1}{l|}{MABP~\cite{li_bring_2024}}                                  & \multicolumn{1}{c|}{Swin-B}                              
& 65.72 & 68.86 & \multicolumn{1}{c|}{62.18}                         & 71.59 & 72.81 & \multicolumn{1}{c|}{-}                         & 62.76      & 64.04     & -     \\
\multicolumn{1}{l|}{CoHD~\cite{luo_hdc_2024}}                                  & \multicolumn{1}{c|}{Swin-B}                              
& 65.17 & 68.42 & \multicolumn{1}{c|}{63.38}                         & 71.85 & 72.67 & \multicolumn{1}{c|}{64.00}                         & 62.63      & 63.60     & 60.37     \\

\rowcolor[HTML]{CBCEFB} 
\multicolumn{1}{l|}{\archname(\textbf{Ours})}                         & \multicolumn{1}{c|}{Swin-B}                              
& \textbf{68.94} & \textbf{74.34} & \multicolumn{1}{c|}{\textbf{79.72}}                         & \textbf{73.22} & \textbf{74.51} & \multicolumn{1}{c|}{\textbf{75.65}}                         & \textbf{63.88}      & \textbf{65.74}     & \textbf{70.72}     \\ \bottomrule
\end{tabular}%
}
\caption{Quantitative comparison of LLM-based, RES, and GRES methods across val, testA, and testB sets of gRefCOCO dataset. \textsuperscript{\textdagger} denotes methods that use external datasets.}
\label{tab:comparison} 
\end{table*}

\section{Experiments}
\label{sec:experiment}
\subsection{Experimental Setup}
\parag{Datasets and Evaluation Metrics.} 
Our experiments are conducted on two primary benchmarks in GRES: gRefCOCO~\cite{liu_gres_2023} and Ref-ZOM~\cite{hu_beyond_2023}, which derive images from COCO dataset~\cite{lin_microsoft_2014, yu_modeling_2016, mao_generation_2016}. The ground-truth instance-level masks are already provided in both gRefCOCO and Ref-ZOM. We assess \archname{} using standard metrics on the gRefCOCO~\cite{liu_gres_2023} and Ref-ZOM~\cite{hu_beyond_2023} datasets. For gRefCOCO, we employ cIoU, gIoU, and N-acc metrics, while for Ref-ZOM, we use mIoU, oIoU, and Acc metrics.

\parag{Implementation Details.} Our model is implemented in PyTorch. Our model uses Swin-Transformer-B~\cite{liu_swin_2021} as the visual encoder and we adopt BERT~\cite{liu_roberta_2019} as our text encoder. The visual encoder is initialized with classification weights pre-trained on ImageNet22K~\cite{russakovsky_imagenet_2015}. We use $K = 9$ blocks of transformer decoder in total. We resize the image input to the resolution of $480 \times 480$ for both training and evaluation. We use the AdamW~\cite{loshchilov_decoupled_2018} optimizer with a batch size of 32. The learning rate is set to $10^{-4}$ and linear decreasing to $10^{-6}$ in 20 epochs. The entire training process takes approximately 24 hours on four NVIDIA A5000 GPUs. More details of our experiential setup are provided in Supplementary Materials.

\subsection{Main Results}
We evaluate {\archname} on the gRefCOCO~\cite{liu_gres_2023} and Ref-ZOM~\cite{hu_beyond_2023} datasets, comparing our performance with state-of-the-art methods in both LLM-based, RES and GRES approaches. Tables \ref{tab:comparison} and \ref{tab:comparison_refzom} summarize the results, demonstrating that {\archname} (highlighted in \colorbox[HTML]{CBCEFB}{lavender}) consistently outperforms existing approaches across all metrics and datasets.

\begin{table}[t]
\centering
\begin{tabular}{@{}lcccc@{}}
\toprule
\multicolumn{1}{l|}{\multirow{2}{*}{\textbf{Method}}} & \multicolumn{1}{c|}{\multirow{2}{*}{\textbf{Backbone}}} & \multicolumn{3}{c}{\textbf{test}} \\ \cmidrule(l){3-5} 
\multicolumn{1}{l|}{}                        & \multicolumn{1}{c|}{}                          & mIoU   & oIoU   & Acc  \\ \midrule
\multicolumn{5}{c}{\textit{LLM-based   Methods}}                                                                                  \\ \midrule
\multicolumn{1}{l|}{LISA-7B\textsuperscript{\textdagger}~\cite{lai_lisa_2024}}                 & \multicolumn{1}{c|}{SAM-ViT-H}                 & 65.39  & 66.41  & 93.39  \\
\multicolumn{1}{l|}{GSVA-7B\textsuperscript{\textdagger}~\cite{xia_gsva_2024}}                 & \multicolumn{1}{c|}{SAM-ViT-H}                 & 68.13  & 68.29  & \textbf{94.59}  \\ \midrule
\multicolumn{5}{c}{\textit{RES   Methods}}                                                                                        \\ \midrule
\multicolumn{1}{l|}{MCN~\cite{luo_multi-task_2020}}                     & \multicolumn{1}{c|}{ResNet-101}                & 54.70  & 55.03  & 75.81  \\
\multicolumn{1}{l|}{CMPC~\cite{liu_cross-modal_2022}}                    & \multicolumn{1}{c|}{DarkNet-53}                & 55.72  & 56.19  & 77.01  \\
\multicolumn{1}{l|}{VLT~\cite{ding_vision-language_2021}}                     & \multicolumn{1}{c|}{ResNet-101}                & 60.43  & 60.21  & 79.16  \\
\multicolumn{1}{l|}{LAVT~\cite{yang_lavt_2022}}                    & \multicolumn{1}{c|}{Swin-B}                    & 64.78  & 64.45  & 83.11  \\ \midrule
\multicolumn{5}{c}{\textit{GRES   Methods}}                                                                                       \\ \midrule
\multicolumn{1}{l|}{DMMI~\cite{hu_beyond_2023}}                    & \multicolumn{1}{c|}{Swin-B}                                      & 68.21  & 68.77  & 87.02  \\
\multicolumn{1}{l|}{CoHD~\cite{luo_hdc_2024}}                     & \multicolumn{1}{c|}{Swin-B}                                         & \underline{69.81}  & \underline{68.99}  & 93.34  \\
\rowcolor[HTML]{CBCEFB} 
\multicolumn{1}{l|}{{\archname(\textbf{Ours})}}                    & \multicolumn{1}{c|}{Swin-B}                                         & \textbf{70.81}     & \textbf{71.13}     & \underline{94.23}     \\ \bottomrule
\end{tabular}%
\caption{Quantitative comparison of LLM-based, RES, and GRES methods on test set of Ref-ZOM dataset. \textsuperscript{\textdagger} denotes using external datasets.}
\label{tab:comparison_refzom} 
\vspace{-5mm}
\end{table}

On the gRefCOCO dataset (Table \ref{tab:comparison}), {\archname} achieves a notable improvement in all metrics across all validation and test splits. Specifically, in the validation set, {\archname} attains the highest scores, with $68.94\%$ in cIoU, $74.34\%$ in gIoU, and $79.72\%$ in N-acc. These scores exceed previous best-performing GRES models, such as LQMFormer \cite{shah_lqmformer_2024}, and significantly outperform RES and LLM-based methods, including LISA-7B \cite{lai_lisa_2024} and GSVA-7B \cite{xia_gsva_2024}. For the Ref-ZOM dataset (Table \ref{tab:comparison_refzom}), {\archname} also achieves state-of-the-art results, with cIoU, gIoU, and N-acc surpassing all previous methods. In particular, {\archname} records a cIoU of $70.8\%$, gIoU of $71.1\%$, and an N-acc of $94.23\%$, demonstrating substantial improvements over both GRES methods like DMMI \cite{hu_beyond_2023} and RES models such as LAVT \cite{yang_lavt_2022}. 
These results confirm that {\archname} not only advances the current state of GRES but also sets new benchmarks for segmentation quality across diverse datasets, solidifying its robustness and adaptability in vision-language tasks.

\definecolor{lightblue}{RGB}{203, 206, 251}

\begin{table}[t]

\end{table}

\begin{table}[t!]
\centering
\resizebox{0.85\linewidth}{!}{%
\begin{tabular}{@{}lccc@{}}
\toprule
\multicolumn{1}{l|}{Setting}                             & cIoU & gIoU & N-acc \\ \midrule
\multicolumn{4}{l}{\textit{(a) Instance Supervision}}                          \\ \midrule
\multicolumn{1}{l|}{None}                                & 63.33    & 66.95    & 70.56     \\
\multicolumn{1}{l|}{Mask2Former}                             & 66.26    & 70.32    & 76.19     \\
\rowcolor[HTML]{CBCEFB} 
\multicolumn{1}{l|}{Mask2Former + POA}                         & \textbf{68.94}    & \textbf{74.34}    & \textbf{79.72}     \\ \midrule

\multicolumn{4}{l}{\textit{(b) Design option of Instance Aggregation}}               \\ \midrule
\multicolumn{1}{l|}{Fixed selection}              & 66.67    & 69.25    & 72.96     \\
\multicolumn{1}{l|}{IA w/o PReLU}              & 68.13    & 72.35    & 78.22     \\
\rowcolor[HTML]{CBCEFB} 
\multicolumn{1}{l|}{IA} & \textbf{68.94}    & \textbf{74.34}    & \textbf{79.72}     \\ \midrule


\multicolumn{4}{l}{\textit{(c) No-target Predictor}}                           \\ \midrule
\rowcolor[HTML]{CBCEFB}
\multicolumn{1}{l|}{Concat($Q_{global}, T_{sen}$)}  & \textbf{68.94}    & \textbf{74.34}    & \textbf{79.72}     \\
\multicolumn{1}{l|}{$Q_{global}$}                   & 67.85    & 71.19    & 75.55     \\
\multicolumn{1}{l|}{$T_{sen}$}& 65.29    & 68.91    & 72.33     \\
\multicolumn{1}{l|}{Average all object queries}                 & 64.01    & 66.29    & 69.23   \\   
\midrule
\multicolumn{4}{l}{\textit{(d) Effect of instance-aware approach}}                          \\ \midrule
\multicolumn{1}{l|}{M2F~\cite{cheng_masked-attention_2022} + InstAlign} & 44.43 & 47.42 & 55.58 \\
\multicolumn{1}{l|}{ReLA~\cite{liu_gres_2023} + InstAlign} & 67.16 & 69.23 & 74.56 \\
\rowcolor{lightblue}
\multicolumn{1}{l|}{\archname(\textbf{Ours})} & \textbf{68.94} & \textbf{74.34} & \textbf{79.72} \\

\midrule

\multicolumn{4}{l}{\textit{(e) Number of queries}}                             \\ \midrule
\multicolumn{1}{l|}{$N = 20$}                                  & 67.64    & 72.67    & 77.25     \\
\multicolumn{1}{l|}{$N = 50$}                                  & 68.56    & 74.23    & \textbf{79.85}     \\
\rowcolor[HTML]{CBCEFB} 
\multicolumn{1}{l|}{$N = 100$}                                 & \textbf{68.94}    & \textbf{74.34}    & 79.72     \\
\multicolumn{1}{l|}{$N = 200$}                                 & 68.01    & 73.24    & 78.12     \\ \bottomrule
\end{tabular}
}%
\caption{\textbf{Ablation studies.} We show: (a)~Instance-aware segmentation framework and our proposed POA loss.
(b)~Instance Aggregation.
(c)~No-target predictor.
(d)~Different Baselines.
(e)~Number of object queries.
}
\label{tab:ablation_full}
\vspace{-5mm}
\end{table}

\subsection{Ablation Studies}
\label{sec:ablation}

We perform several ablation studies to evaluate the effectiveness of the key components in our {\archname} model on gRefCOCO~\cite{liu_gres_2023} dataset. The results are listed in the Table \ref{tab:ablation_full}. We highlight our default configuration with \colorbox[HTML]{CBCEFB}{lavender}. Due to the space limit, we provide more analysis of the ablation study in Supplementary.

\parag{Instance Supervision} 
We examine the effect of different instance supervision methods in Table \ref{tab:ablation_full}(a). Without instance supervision, the model is trained only with the merged mask loss $\mathcal{L}_{merged}$ and no-target loss $\mathcal{L}_{nt}$ (e.g. ReLA~\cite{liu_gres_2023}), achieving only 63.33\% cIoU, 66.95\% gIoU, and 70.56\% N-acc. Adopting standard Mask2Former supervision $\mathcal{L}_{inst}$(w/o $\mathcal{L}_{phrase}$) improves cIoU to 66.26\%, indicating that instance-level learning benefits segmentation accuracy. However, incorporating our Phrase-Object Alignment (POA) $\mathcal{L}_{phrase}$ into the instance supervision $\mathcal{L}_{inst}$ further boosts performance to 68.94\% cIoU, 74.34\% gIoU, and 79.72\% N-acc, demonstrating that explicitly linking object queries to phrases enhances segmentation quality.

\parag{Design options of Instance Aggregation}
As shown in Table \ref{tab:ablation_full}(b), directly selecting high-confidence instances for the final mask achieves 66.67\% cIoU and 72.96\% N-acc. However, by incorporating our IA, we achieve an additional +2.0\% cIoU and +6.0\% N-acc. The IA with PReLU setting achieves the highest performance. This indicates that adaptive aggregation and PReLU activation effectively enhance the merging process and mitigate the risk of missing relevant instances or including irrelevant ones, leading to more precise segmentation.


\parag{No-target Predictor} 
We analyze the no-target predictor configurations in Table~\ref{tab:ablation_full}(c), comparing four settings: (1) $Q_{nt}$, which is concatenation of $Q_{global}$ and $T_{sen}$, (2) using only $Q_{global}$, (3) using only $T_{sen}$, and (4) averaging object queries. Averaging all object queries performs the worst (64.01\% cIoU). Since our model is instance-aware, some queries correspond to high-confidence objects, while others represent background or irrelevant regions. Averaging ignores this distinction, making it harder to correctly classify no-target cases. Our results show that both $Q_{global}$ and $T_{sen}$ improve performance and the combination of both yields the best performance, achieving 68.94\% cIoU, 74.34\% gIoU, and 79.72\% N-acc. This confirms that combining global instance features with sentence-level text embeddings is effective for robust no-target detection.


\parag{Effect of instance-aware approach}
To further validate the effectiveness of our proposed method of instance-aware approach, in Table \ref{tab:ablation_full}(d), we apply our instance-aware design for Mask2Former~\cite{cheng_masked-attention_2022} and ReLA~\cite{liu_gres_2023}. By utilizing the Mask2Former directly without text information guidance, the model randomly predict instances in the image, achieving 44.43\% in cIoU. Incorporating ReLA~\cite{liu_gres_2023} into our design achieves the result to 67.17\% in cIoU. These findings demonstrate that our instance-aware approach significantly improves the ability to handle complex scenarios in GRES.

\parag{Number of queries}
We investigate the impact of the number of queries in Table \ref{tab:ablation_full}(e), testing values of $N \in \{20, 50, 100, 200\}$. Our results show that $N = 100$ provides the best balance, achieving the highest cIoU, gIoU, and N-acc scores. Increasing the number of queries beyond 100 does not yield further improvement, while fewer queries lead to a drop in performance, suggesting that $100$ queries provide optimal candidate instance coverage.

\begin{figure*}[h]
    \centering
    \includegraphics[width=\textwidth,trim=0 20 0 0]{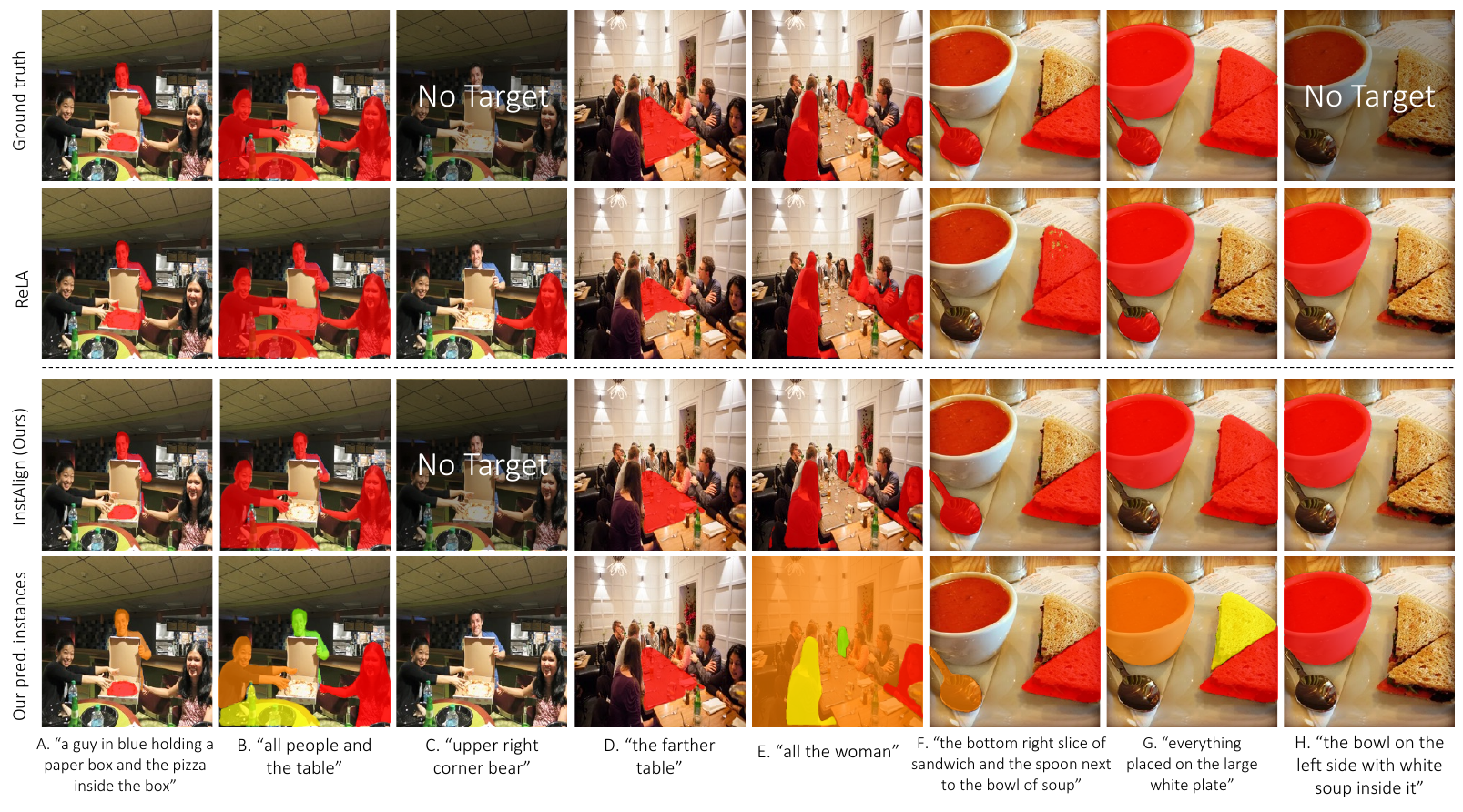}
    \caption{\textbf{Qualitative results on gRefCOCO dataset}. We compare our results with the previous state-of-the-art method ReLA\cite{liu_gres_2023}. We also show our predicted instances with high relevance scores $(\geq0.5)$ in the last row.}
    \label{fig:visualization}
\end{figure*}
\FloatBarrier

\subsection{Visualization}
\label{sec:visualization}
To demonstrate the effectiveness of \archname~in handling complex referring expressions, we present qualitative visualizations comparing our method with the previous state-of-the-art model, ReLA, in Figure \ref{fig:visualization}. In the last row, we show the high-scoring relevant instances predicted by \archname.

ReLA, lacking object-level reasoning, frequently over-segments (columns A, B, C, H), under-segments (columns D, G), or exhibits both issues (columns E, F). In contrast, \archname~explicitly identifies relevant instances and effectively combines them into an accurate segmentation map. In column E, for the expression "all the women," \archname~ correctly segments three individual women at the instance level but includes an extra whole-image prediction, likely influenced by the word "all" in a specific query. However, our IA effectively down-weights this broad instance, integrating relevant lower-scoring predictions to produce a refined segmentation for complex expressions. We provide more visualization and failure cases in Supplementary Materials.


\section{Limitations}
\label{sec:limitation}
Our method is not without limitations. Despite its strong performance, \archname{} faces challenges in interpreting hierarchical or compositional relationships between objects and their attributes, particularly in no-target scenarios. Consequently, the model may fail when expressions include attributes partially describing the target object while introducing conflicting information. For example, in the last column of Figure~\ref{fig:visualization}, \archname{} correctly segments ``the bowl on the left side" but is distracted by the additional attribute ``with the white soup inside it," resulting in inaccurate segmentation.

\section{Conclusion}
We presented \archname{}, a novel instance-aware approach for Generalized Referring Expression Segmentation (GRES). Unlike prior methods that collapse segmentation into a binary foreground mask, \archname{} treats GRES as an instance-level reasoning task. It introduces Phrase-Object Alignment (POA) to explicitly ground object queries in specific parts of the input expression, and Instance Aggregation to selectively merge relevant instances into the final prediction. Through extensive experiments on gRefCOCO and Ref-ZOM, \archname{} consistently outperforms both GRES and LLM-based models, setting a new state of the art. These results highlight the benefits of structured, instance-level grounding for both segmentation accuracy and interpretability. We believe this framework opens new directions for multimodal tasks that demand fine-grained vision-language reasoning.

\section*{Acknoledgements}
This work was supported by Electronics and Telecommunications Research Institute (ETRI) grants funded by the Korean government [24ZR1100, A Study of Hyper-Connected Thinking Internet Technology by autonomous connecting, controlling and evolving ways]. This work was also supported by the Institute of Information \& communications Technology Planning\&Evaluation (IITP) grant funded by the Korea government (MSIT) [No. RS-2024-00336738, Development of Complex Task Planning Technologies for Autonomous Agents, 100\%].

{
    \small
    \bibliographystyle{ieeenat_fullname}
    \bibliography{main}
}

\clearpage
\maketitlesupplementarysingle
\section{Details about Datasets and Metrics}
\subsection{gRefCOCO}
Proposed by Liu \etal~\cite{liu_gres_2023}, the gRefCOCO dataset consists of 19,994 images of 60,287 distinct instances described by 278,232 language expressions. These annotations include 80,022 multi-target and 32,202 no-target samples. 

\noindent For evaluation on gRefCOCO dataset, we use the following metrics:
\begin{itemize}
    \item \textbf{cIoU} (cumulative Intersection over Union): Measures the total pixel intersection over the total pixel union across the validation split, offering a holistic view of segmentation performance.
    \item \textbf{gIoU} (generalized Intersection over Union): Averages the per-image IoU across all samples to evaluate segmentation precision at the image level. For no-target samples, the per-image IoU is 1 if we predict correctly a no-target sample and 0 in the other cases.
    \item \textbf{N-acc} (no-target accuracy): Quantifies the accuracy of no-target identification, i.e., how well the model identifies cases where no target is present.
\end{itemize}

\subsection{Ref-ZOM}
Proposed by Hu\etal~\cite{hu_beyond_2023}, the Ref-ZOM is similar to gRefCOCO, introducing the one-to-one, one-to-many, and one-to-zero referring expressions. These cases correspond to the single-target, multi-target, and no-target samples in gRefCOCO, respectively. Ref-ZOM dataset consists of 55,078 images of 74,942 annotated instances, which includes 41,842 annotated objects under one-to-many settings, 11,937 one-to-zero samples, and 42,421 one-to-one objects. 

For Ref-ZOM, gIoU and cIoU are substituted to the equivalent metrics in RES, mIoU and oIoU. However, different from gRefCOCO, mIoU and gIoU only count for one-to-one and one-to-many samples. For one-to-zero samples,  we use the accuracy (acc) metric, which measures the classification performance on empty-target expressions.

\subsection{Instances ground truth}
While the metrics to measure the performance for GRES task are based on the binary ground-truth mask (that combine all instances together), both gRefCOCO~\cite{liu_gres_2023} and Ref-ZOM~\cite{hu_beyond_2023} datasets provide instance mask annotations for each expression. We utilize these annotations for our instance supervision. This supervision is not external data but comes directly from the gRefCOCO and Ref-ZOM datasets, where instance-level annotations are already provided and accessible to all methods. What distinguishes our approach is that we explicitly model and supervise individual object instances, which previous methods have not done, largely because they lack the instance-aware design needed to make use of this information. Our work aims to fill that gap. We also note that other recent methods have started leveraging richer signals as well—for example, CoHD~\cite{luo_hdc_2024} uses instance-level information for object counting, and LLM-based models~\cite{lai_lisa_2024, xia_gsva_2024} often rely on large-scale external datasets. While our supervision setup is different from earlier works, we believe it’s a principled use of existing data more effectively.

\section{Implementation Details}
Our model is optimized using AdamW~\cite{loshchilov_decoupled_2018} optimizer with the initial learning rate of $10^{-4}$ and linear decreasing to $10^{-6}$ after 20 epochs. Following Mask2Former, , the coefficients of instance supervision are set as $\lambda_{score} = 2, \lambda_{mask} = 5$.$\lambda_{phrase} = 1$. Following ReLA~\cite{liu_gres_2023}, we set $\lambda_{merged} = 1 \text{ and } \lambda_{nt} = 0.1$. We set $\lambda_{phrase} = 1$ and $\lambda_{inst} = 1$ as default hyperparameters.
BERT-base-uncased~\cite{liu_roberta_2019} is used as the Text Encoder to extract language features and the Pixel Decoder comprises 6 layers of Deformable Transformer, following Mask2Former~\cite{cheng_masked-attention_2022}.

\section{Analysis}
\subsection{Phrase-Object Alignment Loss}
The value of $\mathcal{L}_{\text{phrase}}$ directly measures the similarity between object queries and their corresponding phrase embeddings in terms of cosine distance. To better understand how closely these features align, we compute the converged values of $\mathcal{L}_{\text{phrase}}$ on the gRefCOCO validation set under two selection strategies:

\begin{itemize}
    \item \textbf{Oracle setting (ground-truth known):}  
    We assume access to ground-truth instance masks and use Hungarian matching to select the top-$M$ predicted object queries—one for each ground-truth instance. This mirrors the matching logic applied during training.
    
    \item \textbf{Natural setting (inference-time behavior):}  
    No ground-truth information is accessed. Instead, we select predicted object queries based on their relevance scores, retaining those with $\hat{p}_i > 0.5$. This simulates the model's natural selection process at inference time.
\end{itemize}

\begin{table}[h]
\centering
\begin{tabular}{c|c|c|c|c}
\toprule
\textbf{$\mathcal{L}_{\text{phrase}}$} & \textbf{Average} & \textbf{Median} & \textbf{Min} & \textbf{Max} \\
\midrule
\textbf{Oracle setting} & 0.194 & 0.135 & 0.02 & 0.852 \\
\textbf{Natural setting} & 0.142 & 0.105 & 0.03 & 0.697 \\
\bottomrule
\end{tabular}
\end{table}

These values show that while the alignment is meaningful, it is far from perfect—confirming that object queries and phrase embeddings do not collapse into identical representations. The non-negligible distances reflect healthy variation across queries and help preserve distinct semantics, which is essential for accurate instance discrimination.

\subsection{Ablation on loss coefficients}
Since we adopt the framework from Mask2Former~\cite{cheng_masked-attention_2022} and the concept of no-target predictor from ~\cite{liu_gres_2023}, we keep their default hyperparameters. We report the ablation study about the instance supervision  $\lambda_{inst}$ and $\lambda_{phrase}$ in the table below. As we can observe, the balance coefficient between them yields the best performance onthe  gRefCOCO validation set.
\begin{table}[h]
    \centering
    \begin{tabular}{c c | c c}
        \toprule
        $\lambda_{inst}$ & $\lambda_{phrase}$ & cIoU & N-acc. \\
        \midrule
        1  & 1  & \textbf{68.94} & \textbf{79.72}   \\
        1  & 2  & 67.63 & 73.38  \\
        1  & 0.5  & 68.45 & 77.23  \\
        2  & 1  & 68.12 & 77.92  \\
        0.5  & 1  & 67.65 & 76.26  \\
        \bottomrule
    \end{tabular}
    \caption{Ablation study on loss coefficients.}
    \label{tab:loss_ratios}
\end{table}

\subsection{Size and performance}
We provide detailed comparisons of performance, model size and FLOPs on table \ref{tab:performance_comparison}. Our model, \archname, achieves the best accuracy (gIoU: 74.51, cIoU: 68.94, N-acc.: 79.72) while maintaining a balanced computational cost (230M parameters, 0.235T FLOPs).

\begin{table*}[h]
    \centering
    \begin{tabular}{l l c c c c c}
        \toprule
        Method & Backbone & Parameters & T-FLOPs & gIoU & cIoU & N-acc. \\
        \midrule
        DMMI  & Swin-B & 341M  & 0.392T & 62.68 & 62.77 & 53.20 \\
        ReLA  & Swin-B & 226M  & 0.131T & 63.60 & 62.42 & 56.37 \\
        CoHD  & Swin-B & 248M  & 0.133T & 68.42 & 65.17 & 63.68 \\ \midrule
    \archname & Swin-B & 230M  & 0.235T & 74.51 & 68.94 & 79.72 \\
        \bottomrule
    \end{tabular}
    \caption{Performance and efficiency comparison with previous SOTA method on gRefCOCO validation set.}
    \label{tab:performance_comparison}
\end{table*}

Our primary reason for using the Swin-B/BERT-base configuration was to ensure a direct and fair comparison with the most relevant GRES methods, including ReLA, LQMFormer, and CoHD, which all adopt the same Swin-B and BERT-base backbones. Using this configuration enables a clear evaluation of the methodological contributions introduced by our approach. To demonstrate generalization across architectures, we also experiment with additional visual and text encoders, including Swin-L and RoBERTa-base. Results are summarized in Table~\ref{tab:backbone_generalization}.

\begin{table}[h]
\centering
\begin{tabular}{c|c|c|c|c}
\toprule
\textbf{Visual Encoder} & \textbf{Text Encoder} & \textbf{cIoU} & \textbf{gIoU} & \textbf{N-acc} \\
\midrule
Swin-B (default) & BERT & 68.94 & 74.34 & 79.72 \\
Swin-B & RoBERTa & 68.72 & 74.51 & 79.63 \\
Swin-L & BERT & \textbf{69.04} & \textbf{74.82} & \textbf{80.25} \\
\bottomrule
\end{tabular}
\caption{Performance comparison across different backbone configurations.}
\label{tab:backbone_generalization}
\end{table}

\subsection{Design Rationale for Instance Aggregation}
Our instance aggregation module builds on the instance-level object queries and their relevance scores, providing a lightweight refinement that determines how these predictions are merged into the final mask and directly influences segmentation accuracy.
As shown in Table 3(b) main, directly selecting high-confidence instances for the final mask achieves 66.67\% cIoU and 72.96\% N-acc, demonstrating that our instance-aware approach enables the model to predict individual instances independently with reasonable accuracy. However, by incorporating our  Instance Aggregation (IA), we achieve an additional +2.0\% cIoU and +6.0\% N-acc, showing that IA’s soft-assignment approach mitigates the risk of missing relevant instances or including irrelevant ones, leading to more precise segmentation.

One key design choice in IA is performing aggregation on the logits (pre-sigmoid values) of instances rather than their probability scores. This is crucial because logits preserve a linear combination space, allowing the model to better maintain the relative importance of each instance. In contrast, probability values are bounded between 0 and 1, which compresses differences between instances, potentially skewing the merging process. By merging logits instead of probabilities, we avoid this distortion and retain finer segmentation details.

One of the key motivations for choosing PReLU activation is to weaken the influence of negative logits, which typically correspond to irrelevant or background regions. By applying PReLU, we allow the model to suppress negative logits more effectively, enhancing the overall merging process by reducing the contribution of less relevant or distracting instances. While this approach improves the results empirically, understanding why weakening negative logits benefits merging requires further investigation, possibly involving an analysis of how instance logits interact during the combination process.
\subsection{Number of Object Queries}

In both gRefCOCO and Ref-ZOM, a single referring expression can correspond to many distinct object instances—up to 18 in the most extreme cases. Because our training uses Hungarian matching for one-to-one supervision, the number of object queries $N$ must be sufficiently large to accommodate all potentially referred instances. Otherwise, some ground-truth objects would remain unmatched.

We also examined settings with a reduced query capacity. Using only $N = 10$ queries leads to a clear performance drop (cIoU 64.15\%), illustrating that insufficient query slots limit the model’s ability to represent all relevant instances. In our final model, we adopt $N = 100$, which provides a safe margin and consistently stable performance.

\begin{table}[h]
\centering
\begin{tabular}{c|cccc|cc}
\toprule
\multirow{2}{*}{\textbf{Dataset}} & \multicolumn{4}{c|}{\textbf{gRefCOCO}} & \multicolumn{2}{c}{\textbf{Ref-ZOM}} \\
\cmidrule(lr){2-5} \cmidrule(lr){6-7}
 & train & val & testA & testB & train & test \\
\midrule
$M$ & 18 & 14 & 16 & 14 & 18 & 13 \\
\bottomrule
\end{tabular}
\caption{Maximum number of referred object instances ($M$) per expression across datasets.}
\label{tab:max_instances}
\end{table}

\section{Grounding DINO + SAM}
To contextualize the difficulty of Generalized Referring Expression Segmentation (GRES), we evaluate a strong detection-segmentation pipeline using Grounding DINO-base for text-conditioned object detection followed by SAM2.1-Hiera-Large for mask extraction. As shown in Table~\ref{tab:grounding_dino_sam}, this pipeline performs reasonably well on single-object RES (RefCOCO-val), but its performance drops significantly on multi-object and no-target cases in gRefCOCO. This highlights the inherent challenge of GRES and the importance of instance-aware reasoning in our approach.

\begin{table}[h]
\centering
\vspace{-6pt}
\begin{tabular}{ccc|c}
\toprule
\multicolumn{3}{c|}{\textbf{gRefCOCO-val}} & \textbf{RefCOCO-val} \\ 
\textbf{cIoU} & \textbf{gIoU} & \textbf{N-acc} & \textbf{oIoU} \\
\midrule
33.2 & 42.0 & 34.9 & 67.0 \\
\bottomrule
\end{tabular}
\caption{Performance of the Grounding DINO-base + SAM2.1-Hiera-Large baseline.}
\label{tab:grounding_dino_sam}
\end{table}

\begin{table*}[ht]
    \centering
    \resizebox{0.97\textwidth}{!}{%
    \begin{tabular}{lccccccccccc}
        \toprule
        \multirow{2}{*}{Method} & \multirow{2}{*}{Backbone} & \multicolumn{3}{c}{RefCOCO} & \multicolumn{3}{c}{RefCOCO+} & \multicolumn{3}{c}{RefCOCOg} \\
        \cmidrule(lr){3-5} \cmidrule(lr){6-8} \cmidrule(lr){9-10}
         &  & val & test A & test B & val & test A & test B & val & test \\
        \midrule
        VLT \cite{ding_vision-language_2021} & Darknet-53 & 65.65 & 68.29 & 62.73 & 55.50 & 59.20 & 49.36 & 52.99 & 56.65 \\
        ReSTR \cite{kim_restr_2022} & ViT-B/16 & 67.22 & 69.30 & 64.45 & 55.78 & 60.44 & 48.27 & - & - \\
        CRIS \cite{wang_cris_2022} & ResNet-101 & 70.47 & 73.18 & 66.10 & 62.27 & 68.08 & 53.68 & 59.87 & 60.36\\
        LAVT \cite{yang_lavt_2022} & Swin-B & 72.73 & 75.82 & 68.79 & 62.14 & 68.38 & 55.10 & 61.24 & 62.09\\
        ReLA \cite{liu_gres_2023} & Swin-B & 73.82 & 76.48 & 70.18 & 66.04 & 71.02 & 57.65 & 65.00 & 65.97\\
        DMMI \cite{hu_beyond_2023} & Swin-B & 74.14 & 77.13 & 70.16 & 63.98 & 69.73 & 57.03 & 61.98 & 63.46\\
        LQMFormer \cite{shah_lqmformer_2024} & Swin-B & 74.16 & 76.82 & 71.04 & 65.91 & 71.84 & 57.59 & 64.73 & 66.04 \\
        CGFormer \cite{tang_contrastive_2023} & Swin-B & 74.75 & 77.30 & 70.64 & 64.54 & 71.00 & 57.14 & 62.51 & 64.68\\
        PolyFormer$^\dagger$ \cite{liu_polyformer_2023} & Swin-B & 74.82 & 76.64 & 71.06 & 67.64 & 72.89 & 59.33 & 67.76 & 69.05\\
        LISA-7B$^\dagger$ \cite{lai_lisa_2024} & SAM-ViT-H & 74.1 & 76.5 & 71.1 & 62.4 & 67.4 & 56.5 & 66.4 & 68.5 \\
        GSVA-7B$^\dagger$ \cite{xia_gsva_2024} & SAM-ViT-H & 77.2 & 78.9 & 73.5 & 65.9 & 69.6 & 59.8 & 72.7 & 73.3 \\
        MagNet$^\ddagger$~\cite{chng_mask_2024} & Swin-B & {76.55} & {78.27} & {72.15} & {68.10} & {{73.64}} & {61.81} & {67.79} & {69.29}\\ 
        CGFormer \cite{tang_contrastive_2023} & Swin-B & 76.93 & 78.70 & 73.32 & 68.56 & 73.76 & 61.72 & 67.57 & 67.83 \\
        Prompt-RIS$^\dagger$ \cite{shang_prompt-driven_2024} & SAM-ViT-B/16 & 76.36 & 80.37 & 72.29 & 67.06 & 73.58 & 58.96 & 64.79 & 67.16 \\
        VATEX$^\ddagger$ \cite{nguyen-truong_vision-aware_2024} & Swin-B & \textbf{81.53} & \textbf{82.75} & \textbf{79.66} & \textbf{74.61} & \textbf{78.75} & \textbf{68.52} & 75.54 & 76.40 \\
        \midrule
        \rowcolor[HTML]{CBCEFB} 
        \archname{}(\textbf{Ours})$^\ddagger$ & Swin-B & {78.05} & {80.02} & {75.03} & {68.32} & {72.85} & {60.15} & {73.47} & {75.73}\\
        \bottomrule
    \end{tabular}%
    }
    \caption{Comparison with SOTA methods in RES task using the oIoU metric. $^\ddagger$ indicates combining the train splits from these 3 datasets with test images removed to prevent data leakage.  $^\dagger$ indicates using additional data beyond RefCOCO, RefCOCO+, and G-Ref.}
    \label{tab:sota-comparison}
\end{table*}

\section{Instance-aware Supervision details}
Our key design focuses on integrating input language expression into a query-based instance segmentation framework and guiding it to identify and segment only instances described in the input text. Here, we adopt Mask2Former as our instance segmentation framework due to its effectiveness and efficiency. 

\subsection{Feature Extraction}
\label{sec:extraction}
To extract the text information, we adopted BERT-based model~\cite{liu_roberta_2019} to embed the input expression into high-level word features $T_0 \in \mathbb{R}^{L \times C}$, where $C$ and $L$ denotes the number of channels and the length of the expression, respectively. We use an encoder-decoder architecture to extract the multi-scale pixel features $\mathcal{V} = \{V_i \in \mathbb{R}^{C \times H_i \times W_i}\}_{i = 1}^{4}$ from the input image. Here, $H_i$ and $W_i$ denote the height and the width of the feature maps in the $i$-th stage, respectively. As in~\cite{cheng_masked-attention_2022}, these visual features $V_i$ are obtained from the Pixel Decoder~\cite{cheng_masked-attention_2022, zhu_deformable_2021}.

\begin{figure}[t!]
    \centering
    \includegraphics[width=\linewidth,trim=0 0 0 0,clip]{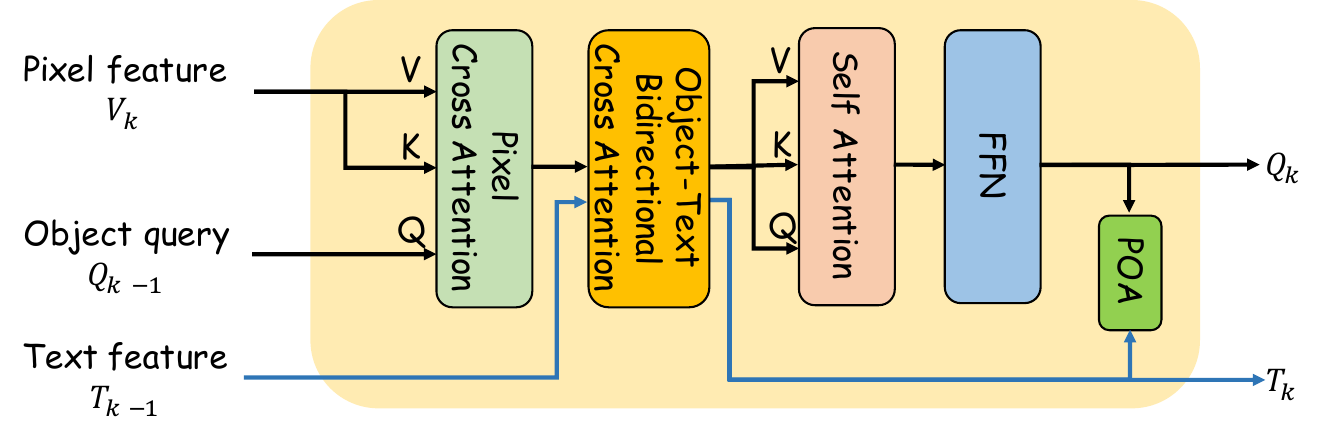}
    \vspace{-5mm}
    \caption{\textbf{Transformer Decoder}. Object query and Text feature are refined by each other and Pixel feature.}
    \label{fig:transformer}
    \vspace{-5mm}
\end{figure}

\subsection{Transformer Decoder}
\label{sec:transformer}
Our transformer decoder is adopted from Mask2Former~\cite{cheng_masked-attention_2022} and ReLA~\cite{liu_gres_2023}. To process an input query, the model operates on a set of $N$ end-to-end trained object queries, alongside extracted text and visual features. Through a sequence of $K$ transformer blocks, illustrated in Fig. \ref{fig:transformer}, it progressively refines these queries by integrating visual and textual features while simultaneously enriching text representations with object-aware information. 

More specifically, the $k$-th transformer block takes the pixel features $V_k \in \mathbb{R}^{H_k \times W_k \times C}$, text features $T_{k - 1} \in \mathbb{R}^{L \times C}$, and object query $Q_{k - 1} \in \mathbb{R}^{N \times C}$ as input and outputs refined object query $Q_{k}$ and  text features $T_{k}$. 
Visual features on specific scales are input to different blocks in a round-robin manner, and where $V_k$ denotes the visual features input to the $k$-th transformer block. The bidirectional object-text cross-attention layer allows both the text features and the object queries to be transformed on the basis of information from both sides. To obtain refined object queries $Q_{k}$, we sequentially pass the object queries $Q_{k - 1}$ through a cross-attention layer with visual features $V_k$, an object-text cross-attention layer with text features $T_{k - 1}$, a self-attention layer and an FFN layer. Simultaneously, the text features $T_{k - 1}$ are passed through the same object-text cross attention layer to produce the refined text features $T_{k}$. We note that this refining text feature is a minor enhancement designed to enrich text representations, and we do not claim contributions from this.

\subsection{Prediction Heads}
Given a refined object query $Q_K \in \mathbb{R}^{N \times C}$, we first predict the probability $\hat{p} \in \mathbb{R}^N$ of these instances being related to the expression input via a simple MLP layer. Similar, $N$ instance masks $\hat{s} \in \mathbb{R}^{\frac{H}{4} \times \frac{W}{4} \times N}$ associated to this object query can be computed by:
\begin{equation}
    \hat{s} = V_4 \cdot Q_K^\top,
\end{equation}
where  $V_4 \in \mathbb{R}^{\frac{H}{4} \times \frac{W}{4} \times C}$ is the pixel features at the highest-resolution scale extracted from the Pixel Decoder (Sec. \ref{sec:extraction})

\subsection{Instance Supervision}
\label{sec:supervision}
The ground truth segmentation \( \mathcal{M}_{gt} \in \mathbb{R}^{H \times W} \) is constructed by combining a set of \( M \) ground-truth instance segments \( s = \{s_i\}_{i = 1}^{M} \). To supervise our predictions effectively, we adopt the instance supervision from Mask2Former~\cite{cheng_masked-attention_2022} to establish a one-to-one correspondence between the predicted instances and the ground-truth instances. Specifically, we find the optimal set of prediction indices \( \omega^* = \{\omega_i\}_{i = 1}^{M} \) that minimizes the following matching cost:
\vspace{-3mm}
\begin{gather}
        \omega^* = \arg \min_{\omega \subseteq [N]} \sum_{i = 1}^{M} \mathcal{L}_{\text{match}}(\omega_i, i),
\end{gather}
where the per-instance matching cost \( \mathcal{L}_{\text{match}} \) for  $i$-th predicted instance and $j$-th ground-truth instance is defined as:
\begin{gather}
    \mathcal{L}_{{match}}(i, j) = \lambda_{{score}}\mathcal{L}_{{score}}(\hat{p}_i, 1)  + \lambda_{{mask}}\mathcal{L}_{{mask}}(\hat{s}_i, s_j)\label{eq:match}.
\end{gather}

Here, \( \mathcal{L}_{{score}} \) is the BCE loss that promotes high confidence for relevant instances, \( \mathcal{L}_{{mask}} \) is the combination of dice loss~\cite{sudre_generalised_2017} and BCE loss to improve mask consistency. The hyper-parameters \( \lambda_{{score}}, \lambda_{{mask}}\) control the balance of each component in the matching cost.

After determining the optimal match, the instance loss $\mathcal{L}_{inst}$ is used to supervise the model for both matched and unmatched samples. The total instance loss is given by:
\begin{equation}
    \mathcal{L}_{inst} = \sum^{M}_{i = 1}{\mathcal{L}_{match}(\omega^*_i, i)} + \sum_{i = 1, i \notin \omega^*}^{N}{\mathcal{L}_{score}(\hat{p}_i, 0)},
\end{equation}
where the first term ensures that for each matched instance, the model minimizes the combined matching loss while the second term penalizes unmatched instances by encouraging their relevance scores \( \hat{p}_i \) to approach zero, indicating irrelevance to the expression. It is important to note that, similar to prior works~\cite{cheng_per-pixel_2021, cheng_masked-attention_2022}, the number of object queries $N$ does not need to equal the actual number of relevant objects $M$.

\section{More quantitative results}
\subsection{Ref-ZOM}
Table 1 showcases the results of our method compared to state-of-the-art methods across one-to-one, one-to-many, and one-to-zero cases in the Ref-ZOM dataset. Our model demonstrates significant improvements across all scenarios. In one-to-one cases, \archname{} surpasses prior SOTA with a margin of 0.93\% and 2.58\% in terms of oIoU and mIoU, respectively, showcasing our precision in identifying individual objects. In one-to-many cases, our model outperforms DMMI~\cite{hu_beyond_2023} with a large margin of 4.47\% and 4.65\% in oIoU and mIoU. For the no-target scenario, the Acc score of $94.23\%$ indicates our model's ability to accurately determine when no valid target is present compared to previous approaches. 

These results highlight the effectiveness of explicitly predicting relevant instances and  instance aggregation, enabling accurate segmentation for complex expressions. 

\begin{table*}[h]
\centering
\resizebox{0.7\textwidth}{!}{%
\begin{tabular}{@{}lccccccc@{}}
\toprule
\multicolumn{1}{l|}{} & \multicolumn{2}{c|}{\textbf{One-to-One}} & \multicolumn{2}{c|}{\textbf{One-to-Many}} & \multicolumn{2}{c|}{\textbf{Overall Targets}} & \multicolumn{1}{c}{\textbf{One-to-Zero}}  \\ \cmidrule(l){2-8} 
\multicolumn{1}{l|}{\multirow{-2}{*}{\textbf{Method}}} & oIoU   & \multicolumn{1}{c|}{mIoU}                         & oIoU  & \multicolumn{1}{c|}{mIoU}                         & oIoU       & \multicolumn{1}{c|}{mIoU}       & Acc     \\ \midrule

\multicolumn{1}{l|}{MCN~\cite{luo_multi-task_2020}} 
& 52.09 & \multicolumn{1}{c|}{53.14} & 58.04  & \multicolumn{1}{c|}{57.21}& 55.05 & \multicolumn{1}{c|}{54.70}    & 75.81     \\
\multicolumn{1}{l|}{CMPC~\cite{liu_cross-modal_2022}}                                                  
& 52.46 & \multicolumn{1}{c|}{52.89} & 60.23 & \multicolumn{1}{c|}{60.27} & 56.19 & \multicolumn{1}{c|}{55.72}        & 77.01     \\
\multicolumn{1}{l|}{VLT~\cite{ding_vision-language_2021}}                                                       
& 59.07 & \multicolumn{1}{c|}{58.96} & 61.42 & \multicolumn{1}{c|}{62.79} & 60.21      & \multicolumn{1}{c|}{60.43}     & 79.26     \\
\multicolumn{1}{l|}{LAVT~\cite{yang_lavt_2022}}                                                           
& 63.21 & \multicolumn{1}{c|}{64.56} & 65.69 & \multicolumn{1}{c|}{65.14} & 64.45      & \multicolumn{1}{c|}{64.78}     & 83.11     \\
\multicolumn{1}{l|}{DMMI~\cite{hu_beyond_2023}}                                                           
& 65.43 & \multicolumn{1}{c|}{66.83} & 72.20 & \multicolumn{1}{c|}{70.44} & 68.77      & \multicolumn{1}{c|}{68.21}     & 87.02     \\ \midrule
\rowcolor[HTML]{CBCEFB} 
\multicolumn{1}{l|}{\archname(\textbf{Ours})}                                          
& \textbf{66.36} & \multicolumn{1}{c|}{\textbf{68.41}}                        & \textbf{76.77} & \multicolumn{1}{c|}{\textbf{75.09}}                         & \textbf{71.13}      & \multicolumn{1}{c|}{\textbf{70.81}}     & \textbf{94.23}     \\ \bottomrule
\end{tabular}%
}
\caption{Quantitative comparison across 3 cases of Ref-ZOM dataset.}
\label{tab:comparison} 
\end{table*}

\subsection{Performance on Traditional RES}
While our primary focus is addressing multi-target scenarios in Generalized Referring Expression Segmentation, we also evaluate our model on the traditional RES task to provide a broader performance perspective. We follow MagNet~\cite{chng_mask_2024} to combine the train splits from 3 datasets RefCOCO, RefCOCO+~\cite{kazemzadeh_referitgame_2014}, and RefCOCOg~\cite{mao_generation_2016} with test images removed to prevent data leakage. As shown in Table 2, while \archname{} is designed for GRES, it also performs competitively on traditional RES benchmarks.

This indicates that our instance-aware reasoning and phrase-object alignment are beneficial even in traditional RES, suggesting broader applicability.

\begin{figure*}[t]
    \centering
    \includegraphics[width=0.9\linewidth]{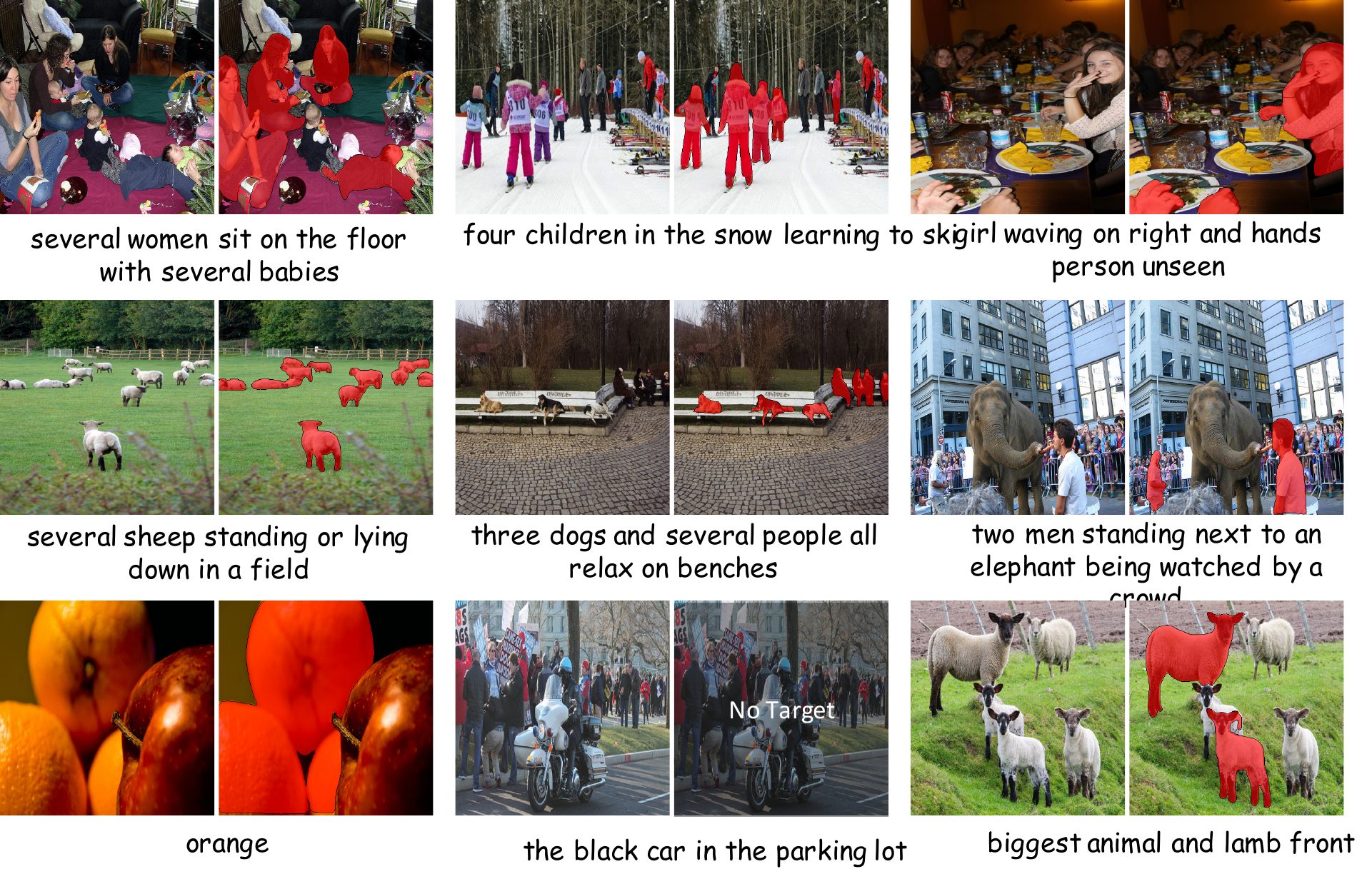}
    \caption{\textbf{Qualitative results on Ref-ZOM dataset.} Our model successfully segments objects based on challenging referring expressions, handling complex multi-object scenarios. Best viewed in color.}
    \label{fig:refzom_visualization}
\end{figure*}

\section{More Qualitative Results}
\label{sec:qualitative_results}

\subsection{Ref-ZOM}
In this subsection, we present qualitative results on the Ref-ZOM dataset, which is designed to test the ability of models to handle complex referring expressions that describe multiple objects or involve intricate spatial relationships. As shown in Fig.~\ref{fig:refzom_visualization}, our model effectively segments the objects based on the input expressions, including multi-object scenarios and fine-grained localization.

\begin{figure*}[t]
    \centering
    \includegraphics[width=0.7\linewidth]{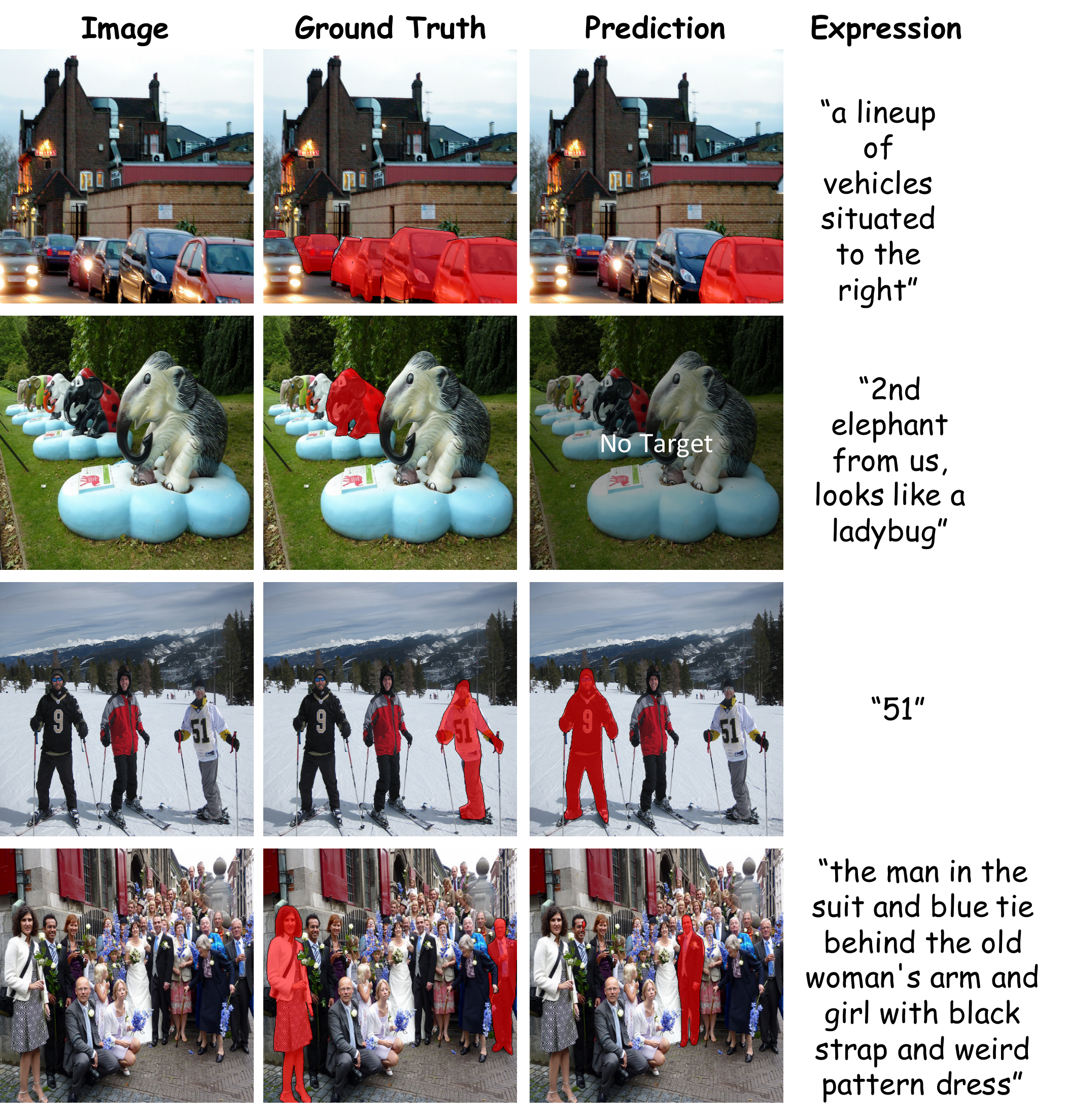}
    \caption{\textbf{Failure case analysis.} We showcase challenging expressions that lead to incorrect or incomplete segmentation results. Some errors occur due to ambiguous expressions (e.g., ``2nd elephant from us''), fine-grained object distinction failures (e.g., ``the man in the suit and blue tie behind the old woman’s arm''), or complex spatial reasoning (e.g., ``a lineup of vehicles situated to the right''). }
    \label{fig:failure_cases}
\end{figure*}

\subsection{Failure Case Analysis}
While our approach achieves promising results in many scenarios, there are certain cases where the model struggles. As shown in Fig.~\ref{fig:failure_cases}, these failure cases primarily arise in three main scenarios: (1) Ambiguous expressions, (2) Fine-grained object distinction failures, and (3) Complex spatial reasoning issues. For instance, expressions like "the second elephant from us" lead to incorrect segmentations because the model has difficulty counting the target from other similar objects in the scene. Similarly, expressions that involve hierarchical spatial relationships, such as "the man in the suit and blue tie behind the old woman's arm,".

A particularly difficult case occurs when segmenting ``51" This requires the model to identify a small, specific region (the number on the shirt) and associate it with the correct individual. Since existing segmentation networks primarily rely on global object shape and texture, fine-grained visual cues like text remain challenging. Future improvements could involve integrating OCR-based feature extraction or multi-level attention mechanisms to better capture textual details within object regions.

Despite these challenges, our model shows strong robustness in the majority of cases, and further research into handling ambiguities and improving spatial reasoning could enhance its performance in these edge cases.


\end{document}